\documentclass[lettersize,journal]{IEEEtran}
\usepackage{amsmath,amsfonts}
\usepackage{algorithmic}
\usepackage{algorithm}
\usepackage{array}
\usepackage[caption=false,font=normalsize,labelfont=sf,textfont=sf]{subfig}
\usepackage{textcomp}
\usepackage{stfloats}
\usepackage{url}
\usepackage{verbatim}
\usepackage{graphicx}
\usepackage{cite}
\usepackage{booktabs}
\usepackage{caption}
\begin{document}
\title{Multi Self-supervised Pre-fine-tuned Transformer Fusion for Better Intelligent Transportation Detection}

\author{
First A. JUWU ZHENG, Second B. JIANGTAO REN
\thanks{This work was supported in part by the Key R\&D projects of GuangDong Province under
Grant 2022B0101070002.\emph{ Corresponging author: Jiangtao Ren.}\\  \indent First A. JUWU ZHENG 
is with the School of Computer Science and Engineering, Sun Yat-Sen University, Guangdong, 510275 
China (e-mail: zhengjuwu29@mail2.sysu.edu.cn).\\ \indent Second B. JIANGTAO REN is with the 
School of Computer Science and Engineering, Sun Yat-Sen University, Guangdong, 510275 
China (e-mail: issrjt@mail.sysu.edu.cn).}
}
\markboth{The paper headers}%
{Shell \MakeLowercase{\textit{et al.}}: Road Disease Detection based on Latent Domain Background Feature Separation and Suppression}
\IEEEpubid{}
\maketitle

\begin{abstract}
    Intelligent transportation system combines advanced information technology to provide intelligent services such as 
    monitoring, detection, and early warning for modern transportation. Intelligent transportation detection is the cornerstone 
    of many intelligent traffic services by identifying task targets through object detection methods. 
    However existing detection methods in intelligent transportation are limited by two aspects. 
    First, there is a difference between the model knowledge pre-trained on large-scale datasets and 
    the knowledge required for target task. Second, most detection models follow the pattern of single-source 
    learning, which limits the learning ability. To address these problems, we propose a \emph{Multi Self-supervised Pre-fine-tuned Transformer Fusion} (MSPTF) network, consisting of two steps: unsupervised 
    pre-fine-tune domain knowledge learning and multi-model fusion target task learning. 
    In the first step, we introduced self-supervised learning methods into transformer model pre-fine-tune which could 
    reduce data costs and alleviate the knowledge gap between pre-trained model and 
    target task. In the second step, we take feature information differences between different model architectures 
    and different pre-fine-tune tasks into account and propose 
    \emph{Multi-model Semantic Consistency Cross-attention Fusion} (MSCCF) network to combine different transformer 
    model features by considering channel semantic consistency and feature vector semantic consistency, 
    which obtain more complete and proper fusion features for detection task. 
    We experimented the proposed method on vehicle recognition dataset and road disease 
    detection dataset and achieved 1.1\%, 5.5\%, 4.2\% improvement compared with baseline 
    and 0.7\%, 1.8\%, 1.7\% compared with sota, which proved the effectiveness of our method.
\end{abstract}

\begin{IEEEkeywords}
    Pre-fine-tune, Broad Learning, Multi-model Fusion, Intellignt Transportation System, Object Detection.
\end{IEEEkeywords}

\section{INTRODUCTION}
    \IEEEPARstart{I}{ntelligent} transportation system is an intelligent service system for transportation based 
    on modern electronic information technology. Through cooperation with advanced information 
    technology such as sensor technology, computer technology, it can reduce transportation 
    pressure, alleviate traffic congestion and provide transportation services. Traditional 
    transportation services are completed by humans, requiring expert knowledge in related fields 
    and a large amount of manpower and material resources. With the widespread application of machine 
    learning in various fields, researchers have successively proposed methods for the field of 
    intelligent transportation system.
\par
    For vehicle classification, \cite{ref221} proposed a channel max pooling scheme and 
    achieve a better vehicle classification result. \cite{ref216} proposed a new double cross-entropy loss 
    function to improve the classification accuracy of transportation vehicles. In the field of vehicle 
    detection, \cite{ref223} used the deep learning model for vehicle damage detection, and demonstrated the 
    learning ability of CNN models with different structures for vehicle damage. \cite{ref201} uses generative 
    adversarial network to generate target domain data from source domain data to achieve vehicle 
    detection enhancement in cross-domain scenarios. On road disease detection, \cite{ref222} adopts 
    multi-scale features and various enhancement methods to improve the performance of disease 
    recognition. \cite{ref208} used a two-stage detection method to detect roadside concrete cracks, and studies 
    the influence of different light and weather conditions on crack detection.
\par
\IEEEpubidadjcol
    However, existing methods for intelligent transportation are 
    usually pre-trained on large datasets and 
    then fine-tune on specific target tasks. These large-scale datasets usually 
    contains common categories like animals and furnitures, which have  
    differences with detection categories in specific fields, such as vehicles or disease categories, 
    and the amount of data in real detection scenarios is limited, which makes it difficult for model to 
    overcome this gap. In \cite{ref224}, 
    the author uses a generator to generate additional synthetic data to improve the robustness of 
    model. In \cite{ref211}, the author proposes a pre-fine-tune method to handle the gap 
    between the knowledge required for sorting and the knowledge extracted by model pre-trained in  
    language task, using additional language understanding tasks to fine-tune 
    the pre-trained model. To address the same problem, \cite{ref206} 
    adopts multiple fine-tuned methods and introduced an intermediate dataset that is closer to the 
    target dataset for initial fine-tune to narrow the difference between model knowledge and the 
    knowledge required for target task. These solutions 
    can be roughly divided into introducing 
    additional datasets that are closer to target task for initial fine-tune and introducing 
    additional network structures, while the former will raise the cost of data acquisition and the latter will make 
    the model more complex. 
\par
    Besides, most of the current machine learning methods are based on single-source learning, which limits the 
    learning ability especially when data is limited, while broad learning can make full use of 
    information from different data, different network structures to make up for the shortcomings 
    of single-source learning \cite{ref230}. \cite{ref228} uses noisy speech and clean reference speech for speech 
    enhancement. \cite{ref214} proposes a 
    cross-modulation strategy to register images from different sensors through dynamic alignment of 
    features, improving the performance of image fusion. 
    \cite{ref219} improves the accuracy of disease diagnosis by concatenating the features of DenseNet\cite{ref212} and 
    ResNet\cite{ref210}. Similar methods include \cite{ref225} and \cite{ref205}, which enhance the learning of target tasks by fusing 
    multiple model features. Broad learning shows great potential in various machine learning tasks, but 
    there is a lack of relevant research in the field of intelligent transportation. The common 
    multi-source methods can be mainly divided into two types, the method for multi-data and the method 
    for multi-structure. In most intelligent transportation scenarios, the data is often directly 
    captured by the camera, and collecting multiple types of data leads to additional cost. 
    In addition, existing multi-structure-based methods perform feature concatenation and 
    discriminator integration, which only model the interaction at a shallow level, and it 
    is difficult to achieve fully fusion between models.
\par
    Based on the above problems, this paper proposes a \emph{Multi Self-supervised Pre-fine-tuned Transformer 
    Fusion} (MSPTF) network. By performing self-supervised pre-fine-tune and multi-model 
    semantic consistency cross-attention fusion on two transformer models, it makes full use of feature 
    information from multiple 
    sources to achieve better real scene detection. Specifically, our method is divided into two 
    steps: self-supervised pre-fine-tune domain knowledge learning and multi-model fusion target task learning. 
    In the first step, we combine self-supervised learning method with model pre-fine-tune. With few training on 
    fine-tune dataset, knowledge gap between pre-trained model and target task is reduced. 
    And the self-supervised learning method 
    allow us to pre-fine-tune on the images containing the target-relevant objects directly, reducing the cost of 
    data collection and eliminates additional annotation overhead. In the second step, we integrate multiple pre-fine-tuned 
    models to make use of knowledge information extracted from different transformer structures and 
    different pre-fine-tune tasks. Specifically we design a \emph{Multi-model Semantic Consistency 
    Cross-attention Fusion} (MSCCF) network to integrate the feature information of Vision Transformer\cite{ref204} into the 
    Swin Transformer\cite{ref220} features. This network considers feature fusion from two 
    aspects. First, correlation between 
    different feature channels of the two model are calculated to achieve channel-wise sementic consistentcy. 
    Second, we calculate the semantic consistency of feature vectors at the same spatial position to weightedly 
    select feature information at different positions. Based on the two consistencies, we obtain the enhanced 
    feature which fully extract information from two models and feed it into the Cascade R-CNN\cite{ref202} 
    framework to train the detection task.
\par
    We trained and verified the model on vehicle model recognition dataset and road disease 
    detection dataset, and experimental results proved the effectiveness of our method. The 
    contributions of this article are as follows:
\begin{itemize}
    \item[$\bullet$]We provide an new pre-fine-tune way to alleviate the knowledge gap between 
    pre-trained model and target task and reduce the pre-fine-tune data cost by combining the self-supervised 
    learning method with model pre-fine-tune.
    \item[$\bullet$]We proposed a multi-model feature fusion network to guide the selective 
    fusion of two transformer features by considering the semantic consistency of feature channels and the 
    semantic consistency of feature vectors at the same spatial position.
    \item[$\bullet$]Based on the above research work, we proposed a Multi Self-supervised Pre-fine-tuned 
    Transformer Fusion network and combined it with the Cascaded R-CNN to improve intelligent transportation 
    detection. 
    Experimental results on vehicle recognition dataset and road disease detection dataset show that our proposed 
    method can effectively improve the detection.
\end{itemize}
\par
    The structure of this paper is as follows: In the second section, we introduce the related work 
    of the research, and propose our method in the third section. In the fourth section, we show the 
    setting and results of the experiment, and summarize the full text in the fifth section.

\section{RELATED WORK}
    In this section, we briefly review the related work of our research.
\subsection{Intelligent Transportation System}
    Intelligent transportation is the product of high-tech development. Through the real-time 
    perception, processing and intelligent decision-making of traffic information and other technical 
    means, it provides intelligent and information-based service modes for urban traffic. Today, with 
    the continuous development of machine learning technology, the combination of machine models and 
    intelligent transportation is also constantly innovating. For example, for vehicle classification, 
    \cite{ref221} proposes that for fine-grained vehicle classification, more discriminative features need to be extracted. 
    Based on this, the author designs a channel max pooling method, which improves the accuracy of fine-grained 
    vehicle classification while reducing the amount of parameters. From the perspective of 
    the loss function, \cite{ref216} points out that the optimization of cross-entropy loss only considers 
    the probability of increasing the sample point belonging to the true value class. The author 
    designs a dual cross-entropy loss function, which further constrains the probability of sample 
    belonging to other classes  and improves the classification 
    effect. For vehicle inspection related fields, \cite{ref223} uses the convolutional network model for vehicle 
    damage detection, showing the application prospects of machine learning models in detecting vehicle 
    damage. Aiming at the problem of domain differences between actual scene data and existing 
    data, \cite{ref201} proposes to use generative adversarial network to build a domain converter, and obtain 
    fake data of the target domain through the converter for training, so as to improve the detection 
    effect in the actual scene. In the road disease 
    detection scene, \cite{ref222} uses EfficientNet for feature extraction, and combines BiFPN to achieve 
    multi-scale feature fusion. \cite{ref218} studies the problem of cross-domain detection and proposed an 
    unsupervised domain adaptation method to learn domain-independent features. To sum up, machine learning methods have important research 
    value in various intelligent transportation scenarios. Based on this prospect, our work studies the 
    application of machine learning models on intelligent transportation.
\subsection{Broad Learning}
    Broad learning was first proposed as a new learning task \cite{ref231}, \cite{ref229} and \cite{ref232}, which mainly fuses 
    multiple large-scale information sources together and mines the fused information in a unified task. 
    The key to broad learning is to fuse information from different sources, and this fusion can be 
    done at different levels. \cite{ref230} summarizes several common extensive learning paradigms, such as raw 
    data level, feature space level, model level and output level. Multi-source in broad learning is an 
    extensive concept, which can refer to different information views, categories, modes, specific sources 
    and domains, such as multi-view, multi-category, multi-domain and multi-modal, etc. For multi-scale 
    learning, \cite{ref203} proposes a multi-scale channel attention module CAM, which fuses multi-scale 
    information by designing convolutions of different sizes. In terms of multi-data source learning, \cite{ref228} takes the noisy speech and clean 
    speech of the same speaker as input and uses 
    the clean reference speech to perform speech enhancement on the noisy speech 
    through the method of feature alignment and fusion. \cite{ref207} proposes an alignment scheme for 
    visual-language multimodal models. The author first uses transformer to extract features of two modalities, 
    and then designs a cross-modal masked reconstruction task to achieve alignment and fusion of different 
    modal features. \cite{ref217} improves the effect of 
    image segmentation by performing layer-by-layer feature fusion on multiple intermediate layers. 
    In addition, on multi-classifier and multi-network structure learning, \cite{ref219}, \cite{ref225} and 
    \cite{ref205} enhance the 
    learning of target tasks by concatenating multiple model features. \cite{ref227} uses the method of ensemble 
    learning for anomaly detection and proposed the P threshold 
    method, which provides a new idea for the integration of multi-model discrimination probabilities.
    \cite{ref213} combines multiple pre-trained models with GAN to provide additional discriminators, 
    which enhances the learning ability of GAN. In summary, broad learning shows great 
     potential in various machine learning tasks, but there is few relevant research in intelligent transportation. 
     In this paper, we attempt to introduce 
     broad learning methods on intelligent transportation tasks.

\section{METHODOLOGY}
\begin{figure*}[!t]
    \centering
    \includegraphics[width=1.0\textwidth]{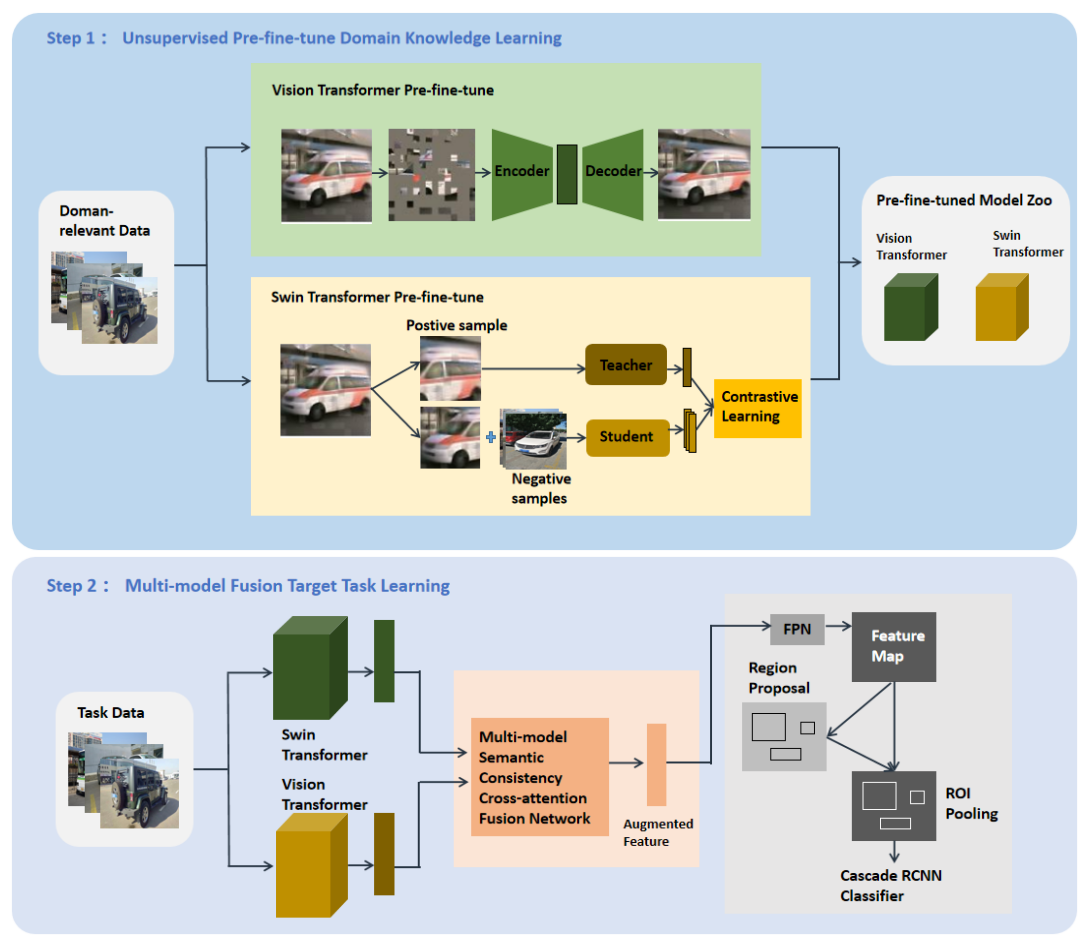}
    \captionsetup{justification=raggedright}
    \caption{The overall architecture of our model, consisting of two steps: unsupervised Pre-fine-tune domain 
    knowledge learning and multi-model fusion target task learning. In step1, we introduce two self-supervised 
    methods to pre-fine-tune two model seperately to alleviate knowledge 
    gap and reduce pre-fine-tune cost. In step2, we extract features propose multi-model sementic consistency cross-attention fusion network 
    to fuse and align features. The obtained augmented features are fed into Cascade R-CNN for training.}
    \label{fig1}
\end{figure*}
    In this chapter, we will introduce the implementation details of the multi-self-supervised 
    pre-fine-tuned transformer fusion network. The overall process is shown in the Fig.\ref{fig1}. 
    The training is divided into two steps, namely self-supervised pre-fine-tune domain 
    knowledge learning and multi-model fusion target task learning. Our work is based on two pre-trained 
    transformer and Cascaded R-CNN \cite{ref202}, in order to alleviate the knowledge gap 
    between pre-trained model and target task, while minimizing the cost of pre-fine-tune, we introduce 
    self-supervised method to pre-fine-tune in the first step. In the second step, we propose a multi-model 
    fusion network to make full use of features extracted from different model and pre-fine-tune tasks. 
    In the following subsections, we will introduce the implementation of each step in detail.

\subsection{Self-supervised Pre-fine-tune Domain Knowledge Learning}
    At present, most of the detection methods in the field of intelligent transportation follow the 
    transfer learning paradigm, in which the model is first pre-trained on 
    large-scale public datasets and then the 
    obtained model is fine-tuned on target tasks. This type of methods ignore the knowledge gap between 
    pre-trained categories and target task categories in specific fields such as vehicle recognition 
    and road disease detection, especially when the target task has limited data, which could influence 
    the fine-tune effect. 
    To solve this 
    problem, the common solution is to pre-fine-tune the pre-trained model by collecting or generating 
    datasets similar to the target task dataset, so that the model can learn on the target task domain 
    to reduce the knowledge gap. 
    But at the same time, pre-fine-tune requires additional data or 
    model structures, which will introduce considerable data collection costs, labeling costs, or more 
    complex model structures.
\par
    Therefore, we combines self-supervised methods with pre-fine-tune, 
    reducing data collection and labeling costs and alleviate knowledge gap through 
    self-supervised pre-fine-tune. As shown in step1 of Fig.\ref{fig1}, we introduce two self-supervised 
    tasks to pre-fine-tune the model. Specifically, we use the masked region modeling in\cite{ref209} to 
    train Vision Transformer\cite{ref204}, and use the contrastive learning method in\cite{ref226} to 
    train Swin Transformer\cite{ref220}. With these two self-supervised method, we can reduce the data collection
    requirement and directly collect images which contain similar objects with target task.
    As shown in step1 of Fig.\ref{fig1}, for 
    vehicle recognition task, we collect images containing random kinds of vehicles to build the pre-fine-tune 
    dataset.
\par
    The mask region modeling pre-fine-tune training is shown in the upper branch of Fig.\ref{fig1}. 
    The input image is divided into tokens and randomly masked, which are input into the 
    Vision Transformer encoder for feature extraction. The extracted feature are processed through 
    the decoder to reconstruct the image. The reconstructed image needs to be close to the real 
    image, thus prompting the encoder to fully extract 
    contextual information to restore the original image. The reconstruction loss evaluated by the difference 
    between the reconstructed area and the original pixel value. The formula is as follows:
    \begin{equation}
        \label{equation_mae}
        \begin{split}
        L_{rec}=\sum (P_{rec}-P_{origin})^2
        \end{split}
    \end{equation}
\par
    where $P_{rec}$ represents the pixel value vector of the reconstructed block and $P_{origin}$ is the pixel value vector of 
    the original image.
\par
    The contrastive learning pre-fine-tune is shown in the lower branch 
    of Fig.\ref{fig1}. We use Swin Transformer as the backbone model. Input image is 
    transformed into two different views through different data augmentations, views from the same picture are regarded as 
    positive samples, and views from different pictures are negative samples. Positive and negative sample 
    pairs are constructed for contrastive learning. The loss function of the model is as follows:
    \begin{equation}
        \label{equation_ssl}
        \begin{split}
        L_{con}=-log\frac{exp(q\cdot k_{+}/\tau)}{\sum\nolimits_{i=0}^K exp(q\cdot k_{i}/\tau)}
        \end{split}
    \end{equation}
\par
    where $q$ is the feature of current view, $k_{+}$ is the feature of the other view  of the same image, 
    $K$ is the total number of positive and negative sample sets, 
    and $\tau$ is the temperature coefficient.
\par
    After self-supervised pre-fine-tune, we obtain a pool of pre-fine-tuned models, and by training 
    with data similar to the target task dataset, the knowledge gap between pre-trained model and 
    target task is reduced.
\subsection{Multi-model Fusion Target Task Learning}
\begin{figure*}[!t]
    \centering
    \includegraphics[width=1.0\textwidth]{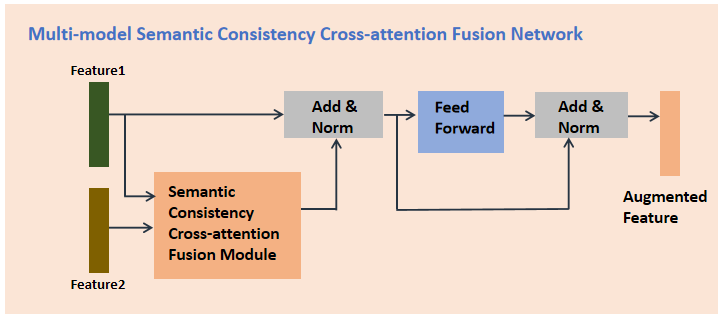}
    \captionsetup{justification=centering}
    \caption{The overall architecture of multi-model semantic consistency cross-attention fusion network, 
    which refers to the cross-attention mechanism. A semantic consistency cross-attention fusion module is 
    designed to achieve better fusion.}
    \label{fig2}
\end{figure*}
    Considering that the model structure and the self-supervised 
    tasks are different, the model can extract features with different semantic information, therefore we fuse 
    different features at this stage to make full use of multi-source information joint learning to 
    improve feature representation, as shown in step2 of Fig.\ref{fig1}. 
    \par
    Based on this, we 
    introduce the broad learning method to fully mine feature information from multiple sources. The 
    most common modes in broad learning are ensemble learning methods and multi-modal fusion methods, 
    but the former often focuses on the integration of decision makers and decision scores, ignoring the 
    interaction of multiple models in the training process, while multi-modal fusion methods are based on 
    multiple data sources paradigm, which is inconsistent with the actual scene of this article.
     Although there are some studies about multi-model learning, most of them are 
     based on feature concatenation or addition, which fail to fully explore and 
     integrate the feature information extracted by multiple models and has limited effect when applied 
     to the target task. Based on the above considerations, we designed a Multi-model 
     Semantic Consistency Cross-attention Fusion network. By modeling the semantic correlation between 
     multiple model feature channels and considering the semantic consistency between corresponding 
     feature points, Vision Transformer feature is selectively integrated into 
     Swin Transformer features to achieve dynamic alignment and fusion of multi-model. 
     The enhanced features are finally input into the Cascaded R-CNN framework.
\par
    The model structure of the multi-model semantic consistency cross-attention fusion network is as 
    shown in Fig.\ref{fig2}. In order to achieve the alignment and fusion of multi-model features, we 
    refer to a similar cross-attention structure. Specifically, we designed a semantic consistency 
    cross-attention fusion module to replace the fusion part in the common cross-attention mechanism for 
    multi-model fusion. Then the features are integrated with Swin Transformer features 
    as residuals and perform normalization, and finally input into the residual feed-forward 
    neural network to obtain the final fusion enhanced features.
\subsection{Semantic Consistency Cross-attention Fusion Module}
\begin{figure*}[!t]
    \centering
    \includegraphics[width=1.0\textwidth]{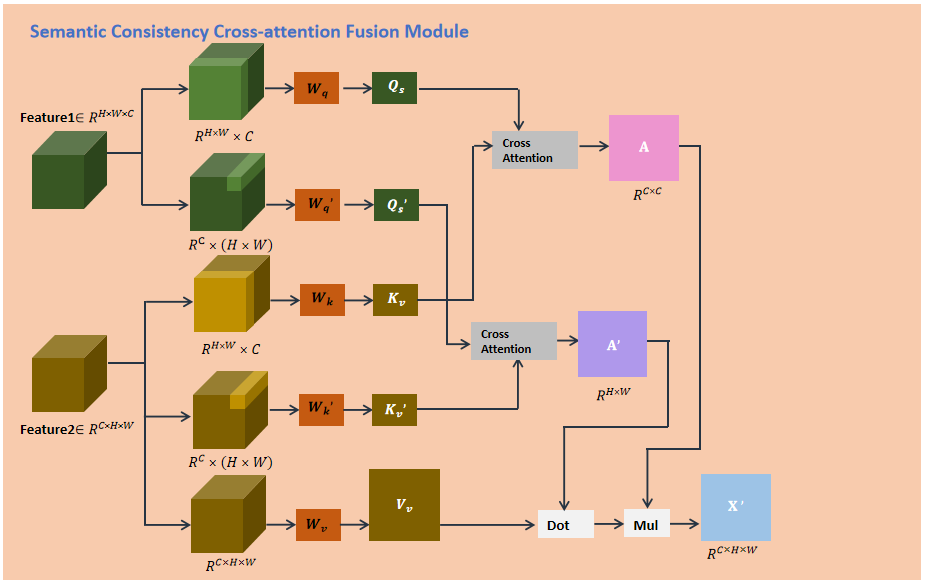}
    \captionsetup{justification=raggedright}
    \caption{The overall architecture of semantic consistency cross-attention fusion module, in which we 
    calculate channel-wise semantic correlation and spatial consistency for feature selection and weighting.}
    \label{fig3}
\end{figure*}
    Cross-attention is often used for multi-modal fusion. The data is usually an image-text pair 
    and the spatial correspondence between image tokens and text tokens is weak, therefore cross-attention between 
    feature vectors at different spatial positions is usually calculated. In our setting, different pre-fine-tuned 
    model features have higher correspondence in spatial dimension while having inconsistent semantic channels, so we 
    design a different semantic consistency 
    cross-attention fusion module to ensure the alignment and fusion of features. 
    \par
    The structure is shown in 
    Fig.\ref{fig3}. Considering that 
    Swin Transformer is pre-trained with contrastive self-supervision, the extracted feature $X_s$ has more 
    discriminative information, and Vision Transformer performs mask region modeling training, therefore 
    feature $X_v$ contains more detailed and complementary information, so we integrate $X_v$ into $X_s$ 
    through cross attention. For the input features $X_s$ and $X_v$, we first input $X_s$ into the convolutional 
    network $W_q$ to obtain the query vector $Q_s$, and input $X_v$ into the convolutional network $W_k$ and $W_v$ 
    to obtain the key vector $K_v$ and the value vector $V_v$, where $Q_s$ and $K_v$ represent the 
    query-key pair used to calculate the correlation between $X_s$ and $X_v$. 
    $V_v$ is obtained from $X_v$, as the filtered information to fuse. As mentioned above, different model features 
    often have inconsistent semantic 
    channels. For the query $Q_s$ and key value $K_v$ which belongs to $R^{C\times H\times W}$, we divide them into $C$ 
    tokens along the channel dimension, which 
    reflects the global semantic features on $C$ channels. Then we calculate the correlation between 
    the $C$ vectors in $Q_s$ and the $C$ vectors in $K_v$. As shown in (\ref{equation_attention_2}), we get the 
    cross attention score map $A\in R^{C\times C}$, which reflects the correlation between each channel in 
    Swin Transformer feature and Vision Transformer feature, therefore the model can weightedly select different 
    channel features of feature $V_v$ to obtain integrated features based on the similarity of channel semantics.
\par
    In addition, considering that each 
    attention score in channel cross attention map represents the overall correlation between the two 
    channels, which take the whole attention map into account with the size of $H\times W$. However, there may be 
    differences in how much information needs to be 
    incorporated at each position, so we add the calculation of semantic consistency of feature
    vectors at same spatial position. For $X_s$ and $X_v$, we use convolutional layers $W_q^{'}$ and $W_k^{'}$ 
    to extract $Q_s^{'}$ and $K_v^{'}$, we employ another two different convolutional layers to make them 
    extract information for comparing the semantic consistency of feature vectors at the same spatial 
    location. So we calculate the similarity of $Q_s^{'}$ and $K_v^{'}$ along the spatial dimension and use $tanh$ as 
    the activation to get the scores. Finally we get $A^{'}\in R^{H\times W}$, which represents the spatial sementic 
    consistency. With the bigger value, we consider that two feature achieve better consistency in current position and 
    assign a smaller weight score, and vice versa.
\par
    For the input features $X_s$ and $X_v$. We obtained the spatial semantic consistency score map $A^{'}$ with 
    size $H\times W$ and the semantic consistency score map $A$ between different channels with size $C\times C$. 
    We perform matrix multiplication between $A$ and $V_v$ to perform channel selection and then multiplied by $A^{'}$ to 
    achieve spatial weighting. The overall fomulas are as follows: 
    \begin{equation}
        \label{equation_attention_1}
        \begin{split}
        Q_s=W_q\cdot X_s , K_v=W_k\cdot X_v , V_v=W_v\cdot X_v
        \end{split}
    \end{equation}
    \begin{equation}
        \label{equation_attention_2}
        \begin{split}
        A(Q_s,K_v)=softmax(\frac{K_v^T Q_s}{\sqrt{d}}),A\in R^{C\times C}
        \end{split}
    \end{equation}
    \begin{equation}
        \label{equation_attention_3}
        \begin{split}
        Q_s^{'}=W_q^{'}\cdot X_s , K_v=W_k^{'}\cdot X_v
        \end{split}
    \end{equation}
    \begin{equation}
        \label{equation_attention_4}
        \begin{split}
        A^{'}(Q_s^{'},K_v^{'})=Tanh(Q_s^{'}-K_v^{'})^2,A^{'}\in R^{H\times W}
        \end{split}
    \end{equation}
    \begin{equation}
        \label{equation_attention_5}
        \begin{split}
        X^{'}=A^{'}(Q_s^{'},K_v^{'})\circ A(Q_s,K_v)V_v
        \end{split}
    \end{equation}
    where $X_s$ and $X_v$ are features from Swin Transformer and Vision Transformer. $W_q$, $W_k$, $W_v$, $W_q^{'}$ 
    and $W_k^{'}$ are convolutional layers. $A$ and $A^{'}$ represent the channel cross attention map and spatial consistency 
    map. $d$ is the length of vectors and $\circ$ is the element-wise dot.

\section{EXPERIMENTS}
    The proposed method is tested on two vehicle recognition datasets and a road disease dataset. 
    In addition, we conduct ablation experiments to verify the performance of each part of the model. 
    In this section, we present the results of individual experiments and discussion.

\subsection{Dataset Setting}
\begin{figure}[!t]
    \includegraphics[width=1.0\columnwidth]{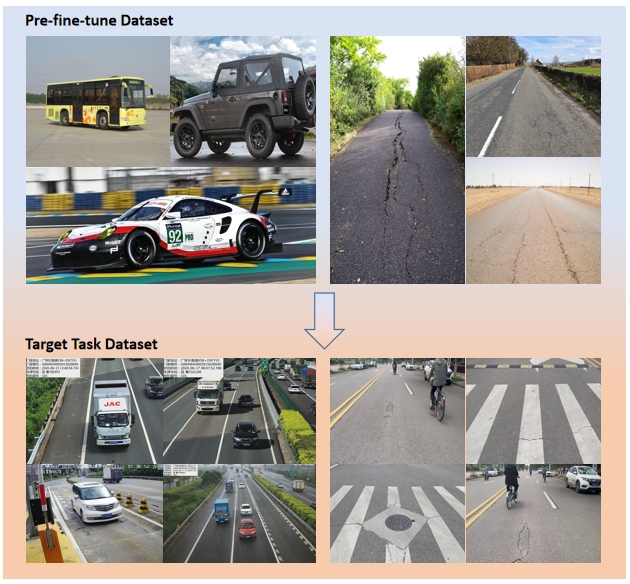}
    \captionsetup{justification=raggedright}
    \caption{Illustration of pre-fine-tune dataset and target task dataset. We collect 
    pre-fine-tune images from internet which contain random types of vehicles and road 
    disease without annotation for self-supervised pre-fine-tune.}
    \label{fig4}
\end{figure}
{\bf{Pre-fine-tune Dataset}} As shown in Fig.\ref{fig4}. We collect images from internet 
to build pre-fine-tune dataset for both vehicle recognition task and road disease detection 
task. We collect 5000 images for vehicle recognition and 3000 images for road disease detection. 
Vehicle pre-fine-tune dataset contains several types of vehicles such as fire engine, suv, racing 
car, etc. For road disease detection, we roughly collect images of damaged pavement for pre-fine-tune. 
It can be seen that pre-fine-tune dataset and the target task dataset are more similar, 
but still maintain certain differences.
\par
{\bf{Vehicle Recognition}} The vehicle recognition dataset contains 
images collected from highway cameras and toll booth cameras, and contains two vehicle types, trucks and 
buses. This dataset is divided into a training set and a test set. The training set contains 1767 
pictures and the test set contains 211 pictures.
\par
{\bf{Vehicle-Dataset}} This dataset collects 3000 images from public road, with a total of 
21 class densely labeled, including bicycles, buses, cars, motorcycles and other common transportation 
categories. We divide the training set and test set with a 4:1 ratio, the training set contains 
2402 pictures and the test set contains 600 pictures.
\par
{\bf{RDD2022}} This data contains pictures of road diseases taken by 
smartphones mounted on motorcycles. The pictures are divided into four different disease categories, 
longitudinal cracks, lateral cracks, alligator cracks and potholes. The number of images is 1977 and we 
devide it by 4 :1, the training set contains 1582 images and the test set contains 395 images.

\subsection{Vehicle-Recognition}
    In order to verify 
    the detection effect of our method, we conducted training and testing on two datasets, the 
    Vehicle-Recognition dataset and the Vehicle-Dataset dataset, and selected four feature extraction models 
    based on the transformer structure as comparison methods. We combined these transformer models 
    with the Cascaded R-CNN detection framework, trained and tested the results on two vehicle recognition 
    datasets.
\par
    The first experiment was conducted on the vehicle recognition dataset. The results are shown in 
    Table \ref{table-vehicle1}. Transformer-SSL and MAE were pre-trained on ImageNet1k based on the self-supervised 
    pre-training method to extract general semantic knowledge. Swin-Transformer and Uniformer 
    were pre-trained on ImageNet1k, perform supervised classification pre-training, extract 
    discriminative semantics of common categories, and then perform target detection task pre-training 
    on the MS COCO dataset. It can be seen from the experimental results that in comparison, the method 
    with additional detection pre-training can achieve better detection results, and our method 
    fully combines the semantic features of the two models to achieve better detection results without 
    object detection pre-training and have the highest mAP. Compared 
    with the baseline Transformer-SSL model, our method achieved an improvement of 1.1\%, and 
    achieved an improvement of 0.7\% compared with Swin Transformer.
\par
    The Second experiment was conducted on Vehicle-Dataset. As shown in Table \ref{table-vehicle2.1} 
    and Table \ref{table-vehicle2.2}, our method also achieved 
    the best mAP, which is 1.8\% higher than other methods. The Vehicle-Dataset 
    dataset contains 21 categories. We selected some categories and displayed the experimental results 
    separately. Table \ref{table-vehicle2.1} shows the AP of major categories in Vehicle-Dataset, each class 
    has thousands of training samples. Table \ref{table-vehicle2.2} show the result of minority class and 
    there are only hundreds of training examples for each class.
     It can be seen from the experimental results that for categories with sufficient samples, 
     Swin-Transformer, Uniformer and our method can achieve better detection results, 
     while for categories with smaller number, our method can achieve better detection results that others in most 
     cases. To a certain extent, this reflects that our model can obtain a more complete feature representation 
     through the method of multi-model feature selection and fusion, thereby improving the detection performance 
     of the model on the minority class.
     \begin{table}
        \begin{center}
        \caption{RESULTS ON VEHICLE-RECOGNITION DATASET}
        \label{table-vehicle1}
        \resizebox{0.50\textwidth}{!}{
        \begin{tabular}{ c | c | c | c  }
        \toprule
        \hline
        Model & CAR & TRUCK & mAP \\
        \hline
        Transformer-SSL\cite{ref226}                          & 85.7 & 89.3 & 87.5 \\
        \hline
        MAE\cite{ref209}                                      & 84.7 & 88.5 & 86.6 \\
        \hline
        Swin Transformer\cite{ref220}              & 86.3 & 89.5 & 87.9 \\
        \hline
        UniFormer\cite{ref215}                     & 85.8 & 89.4 & 87.6 \\
        \hline
        {\bf{ours}}                               & {\bf{87.6}} & {\bf{89.6}} & {\bf{88.6}} \\ 
        \hline
        \bottomrule
        \end{tabular}
        }
        \end{center}
        \end{table}

\begin{table}
    \begin{center}
    \caption{RESULTS ON VEHICLE-DATASET, CLASS TYPE 1}
    \label{table-vehicle2.1}
    \resizebox{0.50\textwidth}{!}{
    \begin{tabular}{ c | c | c | c | c | c | c | c  }
    \toprule
    \hline
    Model & car & rickshaw & bus & three wheels & motobike & truck & mAP(all classes) \\
    \hline
    Transformer-SSL\cite{ref226}                          & 67.1 & 49.9 & 61.2 & 68.7 & 48.7 & 55.4 & 43.6\\
    \hline
    MAE\cite{ref209}                                       & 60.3 & 43.6 & 50.0 & 58.2 & 43.0 & 56.7 & 33.9\\
    \hline
    Swin Transformer\cite{ref220}             & 68.5 & 51.7 & \bf{69.0} & 70.0 & 58.4 & \bf{65.6} & 45.7\\
    \hline
    UniFormer\cite{ref215}                    & 68.6 & \bf{58.8} & 68.7 & \bf{77.3} & \bf{58.8} & 64.6 & 47.3\\
    \hline
    {\bf{ours}}                              & \bf{74.4} & 58.6 & 68.9 & 76.7 & 57.0 & 63.3 & \bf{49.1}\\
    \hline
    \bottomrule
    \end{tabular}
    }
    \end{center}
    \end{table}

    \begin{table}
        \begin{center}
        \caption{RESULTS ON VEHICLE-DATASET, CLASS TYPE 2}
        \label{table-vehicle2.2}
        \resizebox{0.50\textwidth}{!}{
        \begin{tabular}{ c | c | c | c | c | c | c | c  }
        \toprule
        \hline
        Model & pickup & minivan & auto rickshaw & human hauler 
        & wheelbarrow & taxi & mAP(all classes) \\
        \hline
        Transformer-SSL\cite{ref226}                           & 42.6 & 38.8 & 50.3 & 36.4 & 18.2 & 71.7 & 43.6\\
        \hline
        MAE\cite{ref209}                                       & 32.6 & 31.1 & 34.8 & 29.8 & 18.2 & 72.1 & 33.9\\
        \hline
        Swin Transformer\cite{ref220}             & \bf{50.3} & 42.9 & 54.9 & 45.4 & 21.6 & 69.8 & 45.7\\
        \hline
        UniFormer\cite{ref215}                    & 46.9 & 41.1 & \bf{56.2} & 36.4 & 25.3 & 71.7 & 47.3\\
        \hline
        {\bf{ours}}                              & 48.0 & \bf{47.9} & \bf{53.7} & 54.9 & \bf{31.6} & \bf{72.7} & \bf{49.1}\\
        \hline
        \bottomrule
        \end{tabular}
        }
        \end{center}
        \end{table}  
           
\subsection{Road Disease Detection}
    In order to verify that our method can also be applied to the detection of other intelligent 
    transportation scenarios, we also conducted the same experiment on road disease detection task. 
    The results are shown in Table \ref{table-disease}. On the D00 and D10 categories, the performance of Uniformer 
    and our method are better,and Swin Transformer and our method have higher AP on the D40 class, 
    and the results of the three methods are close on the D20 category. The final experimental results 
    show that our method has achieved 1.7\% improvement in mAP, indicating that our 
    method is not limited to common detection objects such as vehicles, but is also effective for 
    special detection targets such as road disease.
        \begin{table}
            \begin{center}
            \caption{RESULTS ON RDD2022 DATASET}
            \label{table-disease}
            \resizebox{0.50\textwidth}{!}{
            \begin{tabular}{ c | c | c | c | c | c   }
            \toprule
            \hline
            Model &D00 & D10 & D20 & D40& mAP \\
            \hline
            Transformer-SSL\cite{ref226}                           & 79.6 & 83.5 & 89.1 & 79.6 & 83.0 \\
            \hline
            MAE\cite{ref209}                                       & 78.8 & 82.1 & 88.9 & 76.2 & 81.5 \\
            \hline
            Swin Transformer\cite{ref220}             & 79.6 & 79.2 & 90.4 & \bf{89.4} & 84.6\\
            \hline
            UniFormer\cite{ref215}                    & 87.6 & \bf{84.6} & 88.5 & 81.2 & 85.5 \\
            \hline
            {\bf{ours}}                              &\bf{87.7}& 82.5 & \bf{90.5} & 88.3 & \bf{87.2} \\
            \hline
            \bottomrule
            \end{tabular}
            }
            \end{center}
            \end{table}

\subsection{Ablation Study}
    The main part of our method is pre-fine-tune and multi-model fusion. In order to verify the effect 
    of each structure, we conducted experiments on vehivle recognition dataset and road disease 
    detection dataset. The results are shown in Table \ref{ablation-1} and Table \ref{ablation-2}. 
    From the overall results of the 
    ablation experiment, it can be seen that both multi-model fusion and pre-fine-tune of domain 
    knowledge can improve the detection effect, among which multi-model fusion has greatly improved 
    the detection of vehicle models and diseases, which are respectively increased by 1.1\% and 4.2\% 
    compared with the baseline, which illustrates the feasibility of using information 
    from multiple models to improve task performance, and also proves that our method can obtain 
    better fusion features. In comparison, the effect of domain pre-fine-tune is smaller, with the 
    respective values increasing by 0.2\% and 0.9\%. Considering that the domain knowledge 
    pre-fine-tune dataset used is relatively small, it may be limited by the size of the dataset.
    \begin{table}
        \begin{center}
        \caption{ABLATION STUDY ON VEHICLE-RECOGNITION}
        \label{ablation-1}
        \resizebox{0.50\textwidth}{!}{
        \begin{tabular}{ c | c | c | c | c | c   }
        \toprule
        \hline
        Model & Multi-model Fusion & Pre-fine-tune  &TRUCK & CAR & map \\
        \hline
        ours                 &x&x        & 85.7 & 89.3 & 87.5\\
        \hline
        ours    &\checkmark&x   & 87.5 & 89.3 & 88.4\\
        \hline
        ours       &\checkmark&\checkmark      & 87.6 & 89.6 & 88.6\\
        \hline
        \bottomrule
        \end{tabular}
        }
        \end{center}
        \end{table}

    \begin{table}
        \begin{center}
        \caption{ABLATION STUDY ON RDD2022 DATASET}
        \label{ablation-2}
        \resizebox{0.50\textwidth}{!}{
        \begin{tabular}{  c | c | c | c | c | c | c | c   }
        \toprule
        \hline
        Model& Multi-model Fusion & Pre-fine-tune  &D00 & D10 & D20 & D40& mAP \\
        \hline
        ours                   &x&x      & 79.6 & 83.5 & 89.1 & 79.6 & 83.0 \\
        \hline
        ours   &\checkmark&x  & 87.2 & 78.6 & 89.9 & 89.7 & 86.3 \\
        \hline
        ours       &\checkmark&\checkmark  & 87.7 & 82.5 & 90.5 & 88.3 & 87.2 \\
        \hline
        \bottomrule
        \end{tabular}
        }
        \end{center}
        \end{table}

\subsection{Domain Pre-fine-tune}
    For the two target task scenarios of this article, vehicle recognition and road disease 
    detection, we use different pre-fine-tuned parameters to initialize the model to analyze the impact 
    of domain knowledge pre-fine-tune on experimental results. As shown in Table \ref{prefinetune-1} 
    and Table \ref{prefinetune-2}, we use 
    three pre-fine-tuned model parameters for the two target tasks, which are self-supervised pre-training 
    on the ImageNet1k dataset, pre-fine-tuned in the vehicle field, and pre-fine-tuned in the road 
    disease field. It can be seen that pre-fine-tuned model using the data of the target 
    field can improve the result of the experiment. Among them, the effect of pre-fine-tune in road 
    disease detection task is better than that in vehicle recognition task. The reason could be that 
    the ImageNet1k 
    dataset contains some vehicle types and lacks knowledge about road diseases, and domain knowledge 
    pre-fine-tune can bring knowledge of road diseases to the model, making the improvement in road 
    disease detection more obvious. Furthermore, using inconsistent domain pre-fine-tune task can lead 
    to performance degradation.
    \begin{table}
        \begin{center}
        \caption{RESULTS OF DIFFERENT PRE-FINE-TUNE METHODS ON VEHICLE-RECOGNITION}
        \label{prefinetune-1}
        \resizebox{0.50\textwidth}{!}{
        \begin{tabular}{ c | c | c | c    }
        \toprule
        \hline
        Pre-fine-tune &TRUCK & CAR & map \\
        \hline
        None                          & 87.5 & 89.3 & 88.4\\
        \hline
        Road Disease Pre-fine-tune    & 86.9 & 89.4 & 87.8\\
        \hline
        Vehicle Pre-fine-tune             & 87.6 & 89.6 & 88.6\\
        \hline
        \bottomrule
        \end{tabular}
        }
        \end{center}
        \end{table}

        \begin{table}
            \begin{center}
            \caption{RESULTS OF DIFFERENT PRE-FINE-TUNE METHODS ON RDD2022 DATASET}
            \label{prefinetune-2}
            \resizebox{0.50\textwidth}{!}{
            \begin{tabular}{ c | c | c | c | c | c   }
            \toprule
            \hline
            Pre-fine-tune &D00 & D10 & D20 & D40& mAP \\
            \hline
            None                          & 87.2 & 78.6 & 89.9 & 89.7 & 86.3 \\
            \hline
            Road Disease Pre-fine-tune    & 87.7 & 82.5 & 90.5 & 88.3 & 87.2 \\
            \hline
            Vehicle Pre-fine-tune         & 85.1 & 76.6 & 89.0 & 87.5 & 84.6 \\
            \hline
            \bottomrule
            \end{tabular}
            }
            \end{center}
            \end{table}

\subsection{Multi-model Fusion visualization}
\begin{figure*}[!t]
    \centering
    \includegraphics[width=1.0\textwidth]{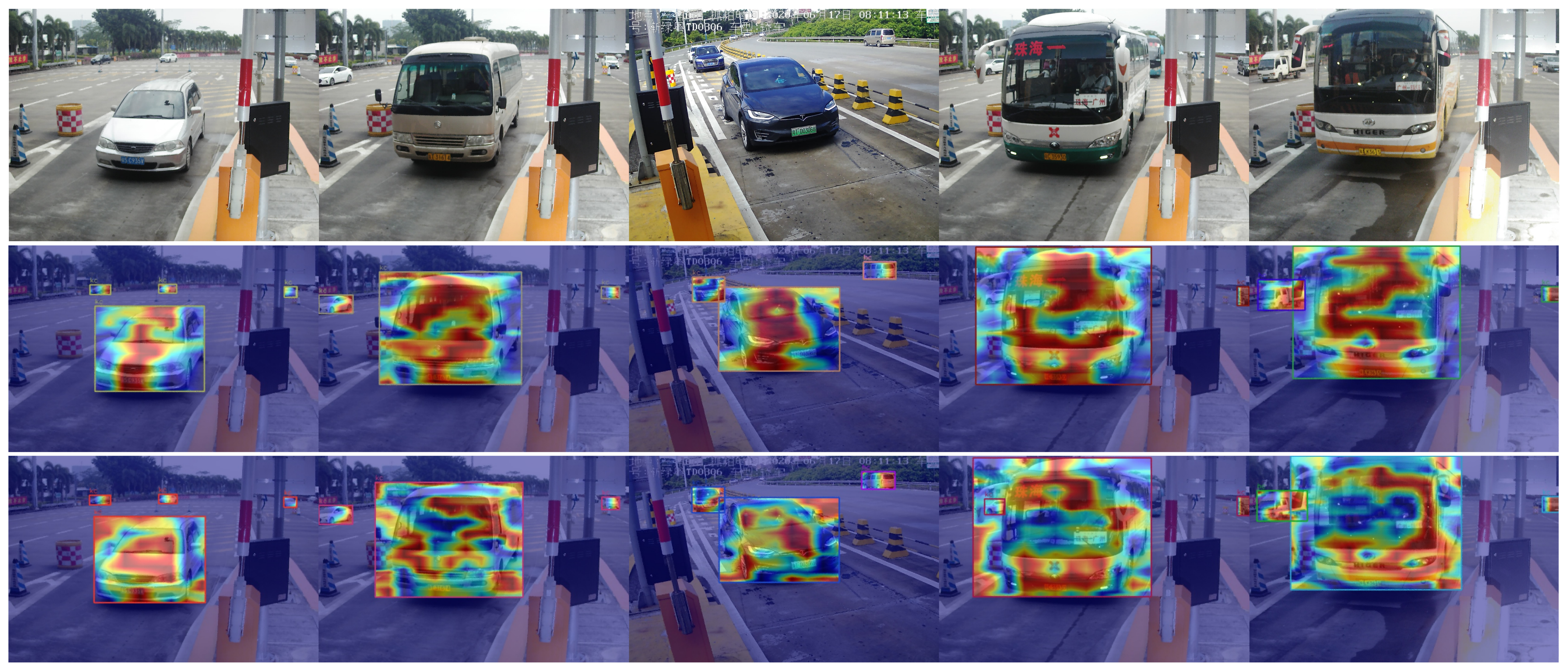}
    \captionsetup{justification=raggedright}
    \caption{Heat map of different model on cars. The rows from top to bottom are 
    input images; visualization of out method, Swin Transformer.}
    \label{fig5}
\end{figure*}

\begin{figure*}[!t]
    \centering
    \includegraphics[width=1.0\textwidth]{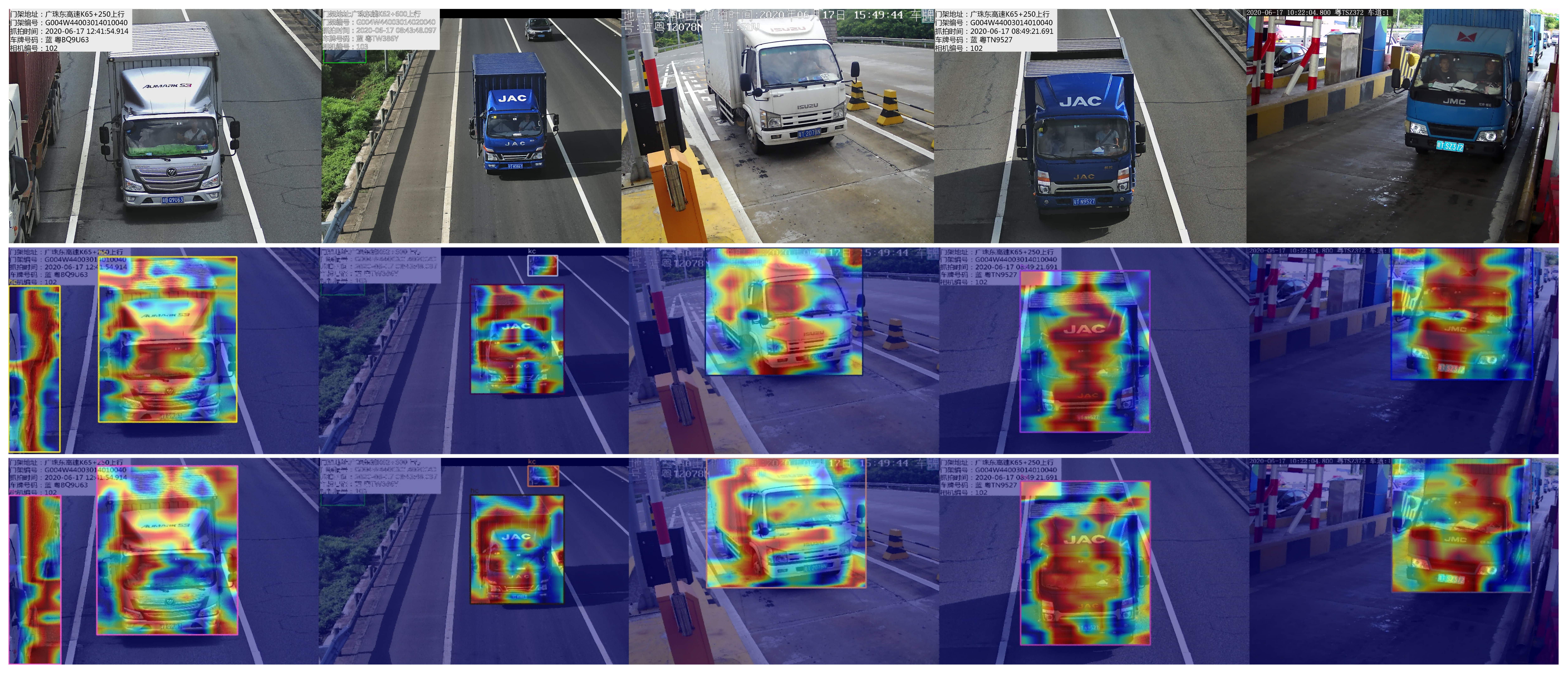}
    \captionsetup{justification=raggedright}
    \caption{Heat map of different model on trucks. The rows from top to bottom are 
    input images; visualization of out method, Swin Transformer.}
    \label{fig6}
\end{figure*}

    We conducted heat map visualization experiments on vehicle recognition dataset, 
    using the GradCAM method to visualize the model's focus areas for trucks 
    and buses. At the same time, we added the visualization results of SwinTransformer for 
    comparison. The experimental results are shown in Fig.\ref{fig5} and 
    Fig.\ref{fig6} and the former visualize attention on car category and latter is for 
    truck. From top to bottom, each row corresponds to original image, visualization results of 
    our model, and visualization results of 
    Swin Transformer. From the visualization results of our model, we can see that the model's focus is 
    usually on the vehicle's front lip, front grille, front hood, windows and other vehicle surface 
    areas, as well as the roof area. From the visualization results of the truck, we can see that the 
    model will pay attention to the compartment and the forehead of the truck, which are relatively 
    more different between the truck and car. In addition, since 
    each vehicle has a relatively similar license plate and headlight appearance, it can be seen that 
    neither method will pay too much attention to these parts which are not suitable for discriminate. 
    Compared with the visualization results of Swin Transformer, our method focuses more on the car 
    body itself in the detection of both vehicle models, and can better explain the model's discriminative 
    learning of the two vehicle types from the perspective of the focus area.

\subsection{Inference Detection}
\begin{figure*}[htbp]
        \begin{minipage}{0.25\textwidth}
            \centering
            \includegraphics[width=\textwidth,height=4\textwidth]{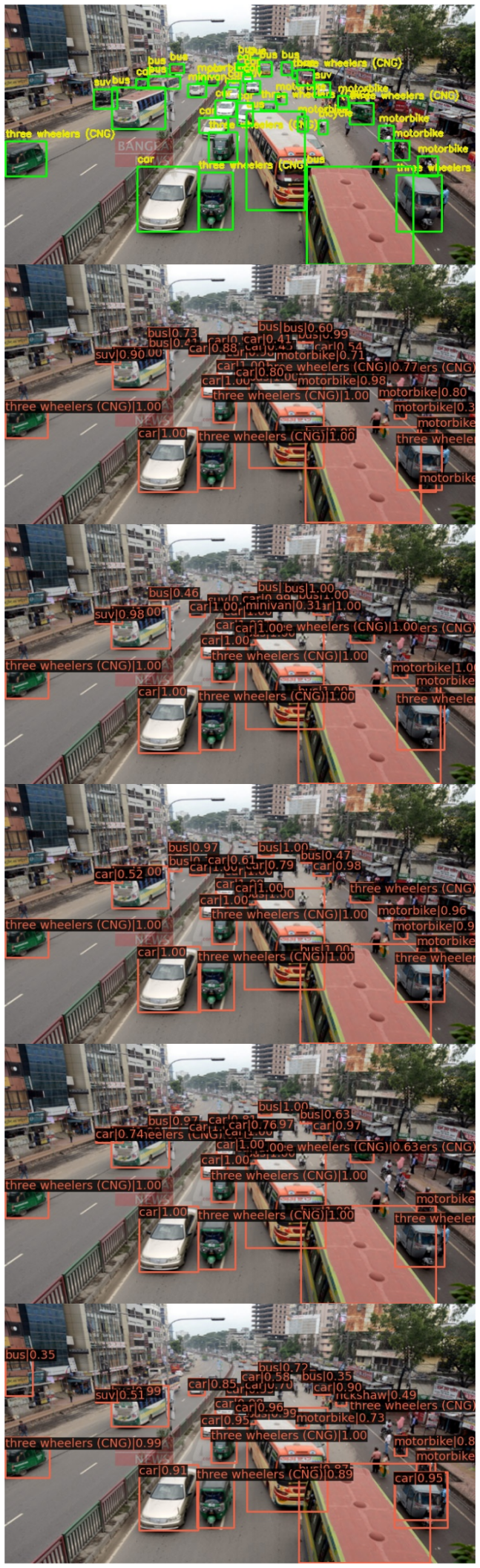}
        \end{minipage}%
        \begin{minipage}{0.25\textwidth}
            \centering
            \includegraphics[width=\textwidth,height=4\textwidth]{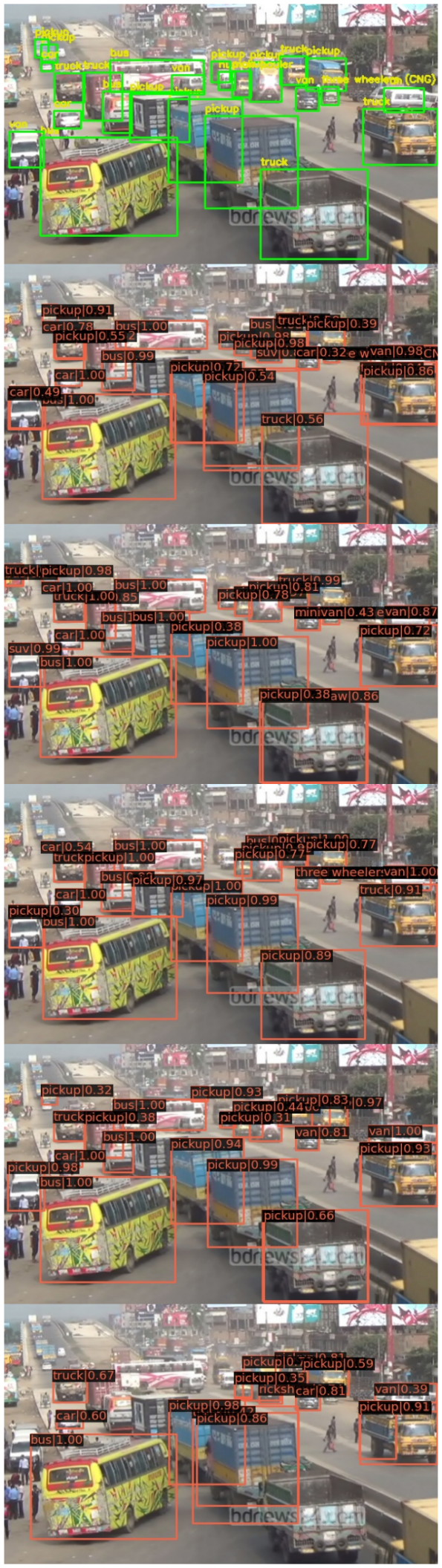}
        \end{minipage}%
        \begin{minipage}{0.25\textwidth}
            \centering
            \includegraphics[width=\textwidth,height=4\textwidth]{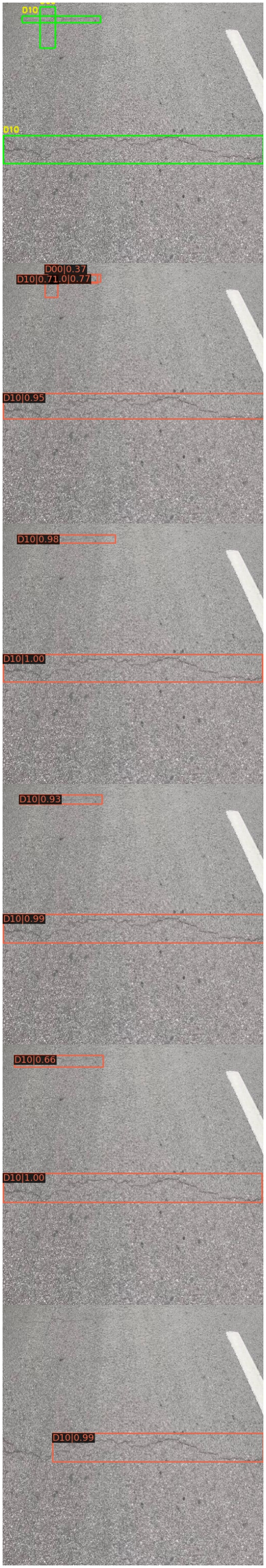}
        \end{minipage}%
        \begin{minipage}{0.25\textwidth}
            \centering
            \includegraphics[width=\textwidth,height=4\textwidth]{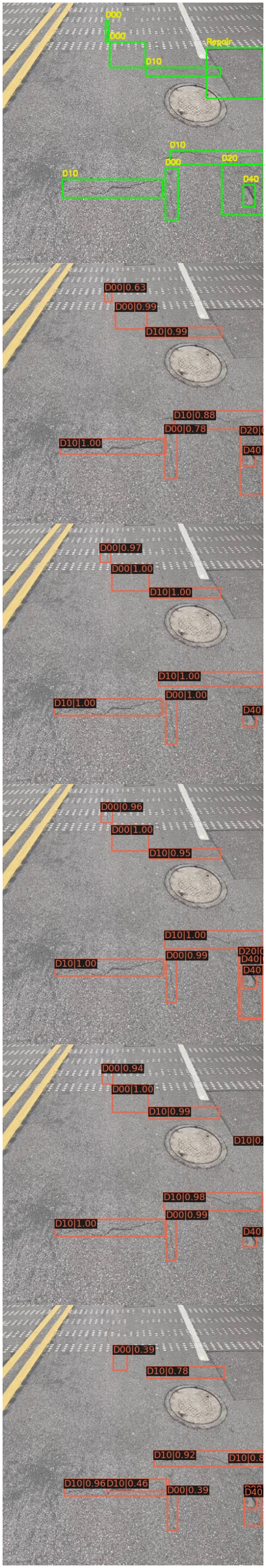}
        \end{minipage}%
        \captionsetup{justification=raggedright}
        \caption{Inference detection results. The rows from top to bottom are 
        GT; detection results of out method, Swin Transformer, Uniformer, 
        Transformer-SSL, MAE.
        }
        \label{fig7}
    \end{figure*}
    Finally, we show the actual detection performance of several methods on the vehicle recognition task 
    and road disease detection task. The detection effect is shown in Fig.\ref{fig7}, each line 
    from top to bottom corresponds to the ground truth bounding box, results of our method, 
    results of Swin Transformer, results of UniFormer, results of Transformer-SSL, and results of MAE. 
    It can be seen from the figure that compared with other models, our method improves the detection 
    accuracy, reduces the missed detection rate, and achieves better detection results in both vehicle 
    recognition and road disease detection tasks.

\section{CONCLUSION}
In this work, we propose a MSPTF network for
intelligent transportation detection task, which could comebines two pre-trained and pre-fine-tune 
transformer to achieve better detection in borad learning manner. Our method includes two step, the first 
step is self-supervised pre-fine-tuned domain knowledge learning, in which we introduce two self-supervised 
methods, masked region modeling and contrasive learning, to pre-fine-tune transformer backbone. Firstly, 
pre-fine-tune helps us alleviate the knowledge gap between pre-trained model and target task. Secondly, we 
reduce the data collection and annotation cost through the combination with self-supervised method. The 
second step is multi-model fusion target task learning, in which we proposed a MSCCF network to make use of features from multi transformer backbone and then 
we combine it with Cascaded R-CNN detection framework to perform training. Specifically, we 
firstly consider the consistency of the channel semantics, and select the different channels of the 
integrated feature by calculating the correlation between different feature channels of the two model. 
Secondly, we calculated the semantic consistency of feature vectors at the same spatial position and used 
it to control the incorporated information at different positions. Finally, we obtain the enhanced fusion 
feature based on two consistencies for subsequent learning. We conducted experiments on vehicle 
recognition datasets and road disease detection dataset and
achieved the best results, proving the effectiveness of our method. In the future, we will 
continue to study more effective and efficient algorithms for better intelligent 
transportation task, and further explore the broad learning method. 
Finally, we hope that our work can stimulate more researches on relevant area.

  \vspace{11pt}
  
  \begin{IEEEbiography}
    [{\includegraphics[width=1in,height=1.25in,clip,keepaspectratio]{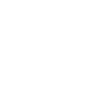}}]
    {JUWU ZHENG}
    received the BE degree from the
    Sun Yat-sen University,
    Guangdong, China, in 2017. He is currently working
    toward the ME degree in the School of Computer 
    Science and Engineering, Sun Yat-sen University. 
    His research interests
    include computer vision and machine learning.
  \end{IEEEbiography}

  \begin{IEEEbiography}
    [{\includegraphics[width=1in,height=1.25in,clip,keepaspectratio]{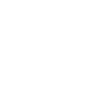}}]
    {JIANGTAO REN}
    received the bachelor's degree in 1998 and the PhD 
    degree in 2003 from Tsinghua University. He is currently an Associate 
    Professor with the School of Computer Science and Engineering, Sun Yat-Sen 
    University. His research interests include data mining and knowledge 
    discovery, machine learning.
  \end{IEEEbiography}

  \vfill


\begin{thebibliography}{99}
  \bibitem{ref201}
  Vinicius F. Arruda,Thiago M. Paix,Alberto F. De Souza, et al., 
  “Cross-Domain Car Detection Using Unsupervised Image-to-Image Translation:From Day to Night,” 
  In \textit{IEEE Transactions on Vehicular Technology}, 
  2019,	arXiv:1907.08719,https://doi.org/10.48550/arXiv.1907.08719.

  \bibitem{ref202}
  Z. Cai, N. Vasconcelos, 
  “Cascade R-CNN: Delving into High Quality Object Detection,” 
  In \textit{2018 IEEE/CVF Conference on Computer Vision and Pattern Recognition}, 
  Salt Lake City, UT, USA, 2018, pp. 6154-6162, doi: 10.1109/CVPR.2018.00644.

  \bibitem{ref203}
  Y. Dai, F. Gieseke, S. Oehmcke, Y. Wu and K. Barnard, 
  “Attentional Feature Fusion,” 
  In \textit{2021 IEEE Winter Conference on Applications of Computer Vision (WACV)}, 
  Waikoloa, HI, USA, 2021, pp. 3559-3568, doi: 10.1109/WACV48630.2021.00360.

  \bibitem{ref204}
  A. Dosovitskiy, L. Beyer, A. Kolesnikov,et al., 
  “An Image is Worth 16x16 Words: Transformers for Image Recognition at Scale,” 
  In \textit{9th International Conference on Learning Representations, ICLR 2021}, 
  https://doi.org/10.48550/arXiv.2010.11929.

  \bibitem{ref205}
  P. Du, E. Li, J. Xia, A. Samat and X. Bai, 
  “Feature and Model Level Fusion of Pretrained CNN for Remote Sensing Scene Classification,” 
  In \textit{IEEE Journal of Selected Topics in Applied Earth Observations and Remote Sensing}, 
  vol. 12, no. 8, pp. 2600-2611, Aug. 2019, doi: 10.1109/JSTARS.2018.2878037.



  \bibitem{ref206}
  M. Gazda, M. Hireš and P. Drotár, 
  “Multiple-Fine-Tuned Convolutional Neural Networks for Parkinson’s Disease Diagnosis From Offline Handwriting,” 
  In \textit{IEEE Transactions on Systems, Man, and Cybernetics: Systems}, 
  vol. 52, no. 1, pp. 78-89, Jan. 2022, doi: 10.1109/TSMC.2020.3048892.

  \bibitem{ref207}
  J. Huang, Y. Li, J. Feng, X. Wu, X. Sun, R. Ji, 
  “Clover: Towards A Unified Video-Language Alignment and Fusion Model,” 
  In \textit{2023 IEEE/CVF Conference on Computer Vision and Pattern Recognition (CVPR)}, 
  2023, Pp.14856-14866,DOI:10.1109/CVPR52729.2023.01427.

  \bibitem{ref208}
  K. Hacıefendio˘glu ,H.B Bas,a˘ga, 
  “Concrete road crack detection using
  deep learning-based faster R-CNN method,” 
  \textit{Iranian Journal of Science and Technology, Transactions of Civil Engineering}, 
  pp.1621–1633,2022.

  \bibitem{ref209}
  K. He, X. Chen, S. Xie, Y. Li, P. Dollar, R. Girshick, 
  “Masked Autoencoders Are Scalable Vision Learners,” 
  In \textit{2022 IEEE/CVF Conference on Computer Vision and Pattern Recognition (CVPR)}, 
  Nov.2017,DOI:https://doi.org/10.48550/arXiv.1711.09409.

  \bibitem{ref210}
  K. He, X. Zhang, S. Ren, J. Sun, 
  “Deep Residual Learning for Image Recognition,” 
  In \textit{CVPR2016}, 
  https://doi.org/10.48550/arXiv.1512.03385.

  \bibitem{ref211}
  X. Hu, S. Yu, C. Xiong, Z. Liu, Z. Liu, G. Yu, 
  “P3 Ranker: Mitigating the Gaps between Pre-training and Ranking Fine-tuning with Prompt-based Learning and Pre-finetuning,” 
  In \textit{Proceedings of the 45th International ACM SIGIR Conference on Research and Development in Information}, 
  Jul. 2022,pp. 1956-1962,https://doi.org/10.1145/3477495.3531786.

  \bibitem{ref212}
  G. Huang, Z. Liu, L.v.d. Maaten, K. Q. Weinberger, 
  “Densely Connected Convolutional Networks,” 
  In \textit{CVPR2017}, 
  https://doi.org/10.48550/arXiv.1608.06993.

  \bibitem{ref213}
  N. Kumari, R. Zhang, E. Shechtman, J. Zhu, 
  “Ensembling Off-the-shelf Models for GAN Training,” 
  In \textit{CVPR 2022 (Oral)}, 
  May 2022,https://doi.org/10.48550/arXiv.2112.09130.

  \bibitem{ref214}
  H. Li, J. Zhao, J. Li , Z. Yu , G. Lu, 
  “Feature dynamic alignment and refinement for infrared-visible image fusion:Translation robust fusion
  Translation robust fusion,” 
  \textit{Information Fusion}, 
  Vol. 95,Issue C,Jul. 2023,pp 26-41,https://doi.org/10.1016/j.inffus.2023.02.011.

  \bibitem{ref215}
  K. Li et al., 
  “UniFormer: Unifying Convolution and Self-Attention for Visual Recognition,” 
  In \textit{IEEE Transactions on Pattern Analysis and Machine Intelligence}, 
  vol. 45, no. 10, pp. 12581-12600, Oct. 2023, doi: 10.1109/TPAMI.2023.3282631.

  \bibitem{ref216}
  X. Li, L. Yu, D. Chang, Z. Ma and J. Cao, 
  “Dual Cross-Entropy Loss for Small-Sample Fine-Grained Vehicle Classification,” 
  In \textit{IEEE Transactions on Vehicular Technology}, 
  vol. 68, no. 5, pp. 4204-4212, May 2019, doi: 10.1109/TVT.2019.2895651.

  \bibitem{ref217}
  Y. Liang , Gu. Qin, M. Sun , J. Qin, J. Yan, Z. Zhang , 
  “Multi-modal interactive attention and dual progressive decoding network for RGB-D/T salient object detection,” 
  \textit{NeuroImage}, 
  Vol. 490,Issue C,Jun. 2022,pp. 132-145,https://doi.org/10.1016/j.neucom.2022.03.029.

  \bibitem{ref218}
  C. Lin, D. Tian, X. Duan, J. Zhou, D. Zhao and D. Cao, 
  “DARDD: Toward Domain Adaptive Road Damage Detection Across Different Countries,” 
  In \textit{IEEE Transactions on Intelligent Transportation Systems}, 
  vol. 24, no. 3, pp. 3091-3103, March 2023, doi:
  10.1109/TITS.2022.3221067.

  \bibitem{ref219}
  M.Liu , F.Li , H. Yan, K. Wang,et al., 
  “A multi-model deep convolutional neural network for automatic
  hippocampus segmentation and classification in Alzheimer’s disease,” 
  \textit{NeuroImage}, 
  vol. 208, pp. 116459, Dec. 2019, DOI: 10.1109/TVT.2019.2899972.

  \bibitem{ref220}
  Z. Liu, Y. Lin,et al., 
  “Swin Transformer: Hierarchical Vision Transformer using Shifted Windows,” 
  In \textit{2021 IEEE/CVF International Conference on Computer Vision (ICCV)}, 
  2021,DOI:10.1109/ICCV48922.2021.00986.

  \bibitem{ref221}
  Z. Ma et al., 
  “Fine-Grained Vehicle Classification With Channel Max Pooling Modified CNNs,” 
  In \textit{IEEE Transactions on Vehicular Technology}, 
  vol. 68, no. 4, pp. 3224-3233, April 2019, doi: 10.1109/TVT.2019.2899972.

  \bibitem{ref222}
  S. Naddaf-Sh,M.M. Naddaf-Sh ,A.R. Kashani, and H. Zargarzadeh, 
  “An Efficient and Scalable Deep Learning Approach for Road Damage
  Detection,” 
  In \textit{2020 IEEE International Conference on Big Data (Big
  Data)}, 
  pp. 5602-5608, 2020.

  \bibitem{ref223}
  K. Patil, M. Kulkarni, A. Sriraman and S. Karande, 
  “Deep Learning Based Car Damage Classification,” 
  In \textit{2017 16th IEEE International Conference on Machine Learning and Applications (ICMLA)}, 
  Cancun, Mexico, 2017, pp. 50-54, doi: 10.1109/ICMLA.2017.0-179.

  \bibitem{ref224}
  R. G. Reddy, V. Yadav, M. A. Sultan, M. Franz, V. Castelli, H. Ji, A. Sil, 
  “Towards Robust Neural Retrieval with Source Domain Synthetic
  Pre-Finetuning,” 
  In \textit{Proceedings of the 29th International Conference on Computational Linguistics}, 
  Oct. 2022,pp.1065-1070,https://doi.org/10.48550/arXiv.2104.07800.

  \bibitem{ref225}
  M. S. Shahabi, A. Shalbaf , A. Maghsoudi , 
  “Prediction of drug response in major depressive disorder using ensemble of transfer learning with convolutional neural network based on EEG,” 
  \textit{Biocybernetics and Biomedical Engineering}, 
  Vol. 41, Issue 3, Sep. 2021, pp. 946-959, https://doi.org/10.1016/j.bbe.2021.06.006.

  \bibitem{ref226}
  Z. Xie, Y. Lin, Z. Yao, Z. Zhang, Q. Dai, Y. Cao , H. Hu, 
  “Self-Supervised Learning with Swin Transformers,” 
  May 2021,DOI:https://doi.org/10.48550/arXiv.2105.04553.

  \bibitem{ref227}
  F. Xue, Z. He, C. Xie, F. Tan, Z. Li, 
  “Boosting Out-of-Distribution Detection with Multiple Pre-trained Models,” 
  In \textit{ICLR 2023}, 
  Feb 2023,https://doi.org/10.48550/arXiv.2212.12720.

  \bibitem{ref228}
  H. Yue,  W.Duo,X. Peng, J.Yang , 
  “Reference-Based Speech Enhancement via Feature Alignment and Fusion Network,” 
  In \textit{Proceedings of the AAAI Conference on Artificial Intelligence}, 
  36(10), 11648-11656. https://doi.org/10.1609/aaai.v36i10.21419.  

  \bibitem{ref229}
  J. Zhang, L. Cui, P. Yu, Y. Lv, 
  “BL-ECD: broad learning based enterprise community detection via hierarchical
  structure fusion,” 
  In \textit{Proceedings of the 2017 ACM on Conference on Information and Knowledge Management}, 
  Nov. 2017,pp. 859-868,https://doi.org/10.1145/3132847.3133026.

  \bibitem{ref230}
  J. Zhang ,Philip S. Yu , 
  “Broad Learning Introduction,” 
  \textit{Broad Learning Through Fusions}, 
  pp. 3-17.

  \bibitem{ref231}
  J. Zhang, C. Xia, C. Zhang, L. Cui, Y. Fu, P. Yu, 
  “BL-MNE: emerging heterogeneous social network embedding
  through broad learning with aligned autoencoder,” 
  In \textit{Proceedings of the 2017 IEEE International Conference on
  Data Mining}, 
  New Orleans, LA, USA, 2017, pp. 605-614, doi: 10.1109/ICDM.2017.70.

  \bibitem{ref232}
  J. Zhu, J. Zhang, L. He, Q. Wu, B. Zhou, C. Zhang, P. Yu, 
  “Broad learning based multi-source collaborative
  recommendation,” 
  In \textit{Proceedings of the 2017 ACM on Conference on Information and Knowledge Management}, 
  Nov. 2017,pp. 1409-1418,https://doi.org/10.1145/3132847.3132976.


  \end{thebibliography}
\end{document}